\documentclass{article}

\PassOptionsToPackage{numbers, compress}{natbib}
\usepackage[dblblindworkshop, final]{neurips_2026}
\workshoptitle{NeurIPS 2026 Workshop on Continual World Models}
\usepackage[utf8]{inputenc} % allow utf-8 input
\usepackage[T1]{fontenc}    % use 8-bit T1 fonts
\usepackage{hyperref}       % hyperlinks
\usepackage{url}            % simple URL typesetting
\usepackage{booktabs}       % professional-quality tables
\usepackage{graphicx}       % result figures
\usepackage{tabularx}       % compact full-width tables
\usepackage{colortbl}       % restrained table highlighting
\usepackage{enumitem}       % compact experiment-note lists
\usepackage{placeins}       % keep floats within their sections
\usepackage{amsmath}        % equations and mathematical notation
\usepackage{amsfonts}       % blackboard math symbols
\usepackage{nicefrac}       % compact symbols for 1/2, etc.
\usepackage{microtype}      % microtypography
\usepackage{xcolor}         % colors

\title{Few-Shot Residual Adaptation of Pretrained Latent World Models}
\author{%
  Jakub Suliga \\
  Karlsruhe Institute of Technology\\
  \texttt{jakub.suliga@student.kit.edu} \\
  \And
  Pankhuri Vanjani \\
  Karlsruhe Institute of Technology\\
  \texttt{pankhuri.vanjani@kit.edu} \\
  \And
  Nisarga Nilavadi \\
  University of Technology Nuremberg\\
  \texttt{nisarga.nilavadi@utn.de} \\
  \AND
  Wolfram Burgard \\
  University of Technology Nuremberg\\
  \texttt{wolfram.burgard@utn.de} \\
  \And
  Rudolf Lioutikov \\
  Karlsruhe Institute of Technology\\
  \texttt{rudolf.lioutikov@kit.edu} \\
}

\begin{document}

\maketitle

\begin{abstract}

Pretrained latent world models can support visual planning and control, but their predictions may fail when physical dynamics change at deployment. We study whether such models can adapt from only a few interactions without modifying their pretrained weights or identifying the changed physical parameters. We introduce a few-shot residual adaptation approach that learns self-supervised corrections from latent prediction errors. The corrections are applied both inside the latent predictor and directly to its predicted latent state, while the pretrained world model remains frozen. We evaluate the approach in simulation and on a real robot under center-of-mass and friction shifts. These changes reduce the success of the frozen model to near zero. Using three interaction episodes for adaptation, our dual-site approach recovers 90\% success under the center-of-mass shift and 100\% under the modified friction on the real robot. Ablations show that correcting both locations outperforms single-site adaptation. Our diagnostics further suggest that much of the learned correction reduces a general prediction mismatch, while successful control also depends on shift-specific information captured during adaptation.
\end{abstract}

\section{Introduction}

World models predict how observations evolve under
robot actions and have been used to support planning and control
\citep{hou2026world, hafner2023mastering, wu2023daydreamer}. Latent world models are particularly promising because they can learn task-relevant representations without predicting every pixel, enabling efficient model-based control from visual observations \citep{assran2025v,zhou2024dino,hansen2024tdmpc2}.
A central challenge is that a model trained under one set of physical conditions may be deployed under another. Changes in center of mass, friction, payload, actuator properties, or contact dynamics can alter the effect of the same action even when the task, visual appearance, and goal remain unchanged~\citep{li2025pin,lyu2025dywa,baumeister2025incremental}.
Existing approaches often address this mismatch by retraining the model \citep{levy2026simulation}, estimating
physical parameters \citep{li2025pin, baumeister2025incremental}, adapting a policy \citep{lyu2025dywa, kumar2021rma}, or randomizing dynamics during training \citep{peng2018sim}. These strategies can be effective, but they may require substantial data, access to the training procedure, a suitable parameterization of the dynamics, or prior knowledge about the deployment shift.

In this work, we study how a pretrained latent world model can be efficiently adapted from a few interactions. We keep the pretrained model frozen and learn lightweight residual corrections from its prediction errors. These corrections are trained self-supervised from observed transitions, requiring neither rewards nor knowledge of the changed physical parameters. We inject them at two points in the latent dynamics pipeline: within the predictor and on its predicted latent state. 
We evaluate this setting in simulation and on a real robot under shifted
center-of-mass and modified friction setting. These changes reduce the success of the
frozen world model to near zero, while adaptation from only three interaction
episodes recovers most of the lost performance. We find that adapting
both inside the predictor and at its output works better than adapting either
location alone. Our diagnostics further show that much of the correction reduces a general prediction mismatch, while the large control gap between general- and
shifted-dynamics adapters suggests that some of the learned correction depends
on the dynamics seen during adaptation.

Our main contributions are:
\begin{enumerate}[leftmargin=*,nosep]
    \item \textbf{Dynamics-shift failure:} We show that physical dynamics shifts can break control while the task and controller remain fixed.
    
    \item \textbf{Few-shot residual adaptation:} We restore control from a few self-supervised interactions while keeping the pretrained world model frozen.
    
    \item \textbf{Dual-site correction:} We show that adapting both inside the predictor and on its output outperforms single-site alternatives.
\end{enumerate}

\section{Related Work}

\paragraph{JEPA-style latent world models}
Several works investigate action-conditioned prediction for visual planning. V-JEPA 2-AC, a joint-embedding predictive architecture, predicts future latent representations conditioned on robot actions~\citep{assran2025v}. 
Similarly, DINO-WM predicts future DINOv2 patch representations from observation and action histories and employs model-predictive control to reach visual goals~\citep{zhou2024dino}. 
Other latent world models, including DreamerV3~\citep{hafner2023mastering}, TD-MPC2~\citep{hansen2024tdmpc2}, and DayDreamer~\citep{wu2023daydreamer}, learn compact predictive representations for control, but generally couple world-model learning with task-specific reward or value prediction. 

\paragraph{Parameter-efficient and test-time adaptation}
Parameter-efficient fine-tuning (PEFT) adapts a pretrained network while keeping most parameters frozen. LoRA uses low-rank weight updates~\citep{hu2021lora}, bottleneck adapters add nonlinear residual branches~\citep{houlsby2019parameter}, and Adapter$^{+}$ scales such a branch after the feed-forward block~\citep{steitz2024adapters}. Other work instead corrects the network output~\citep{zeng2020tossingbot}. Test-time adaptation updates deployed models from unlabeled data using self-supervised or entropy-based objectives~\citep{sun2020test,wang2020tent,hansen2020self}. We combine these ideas by learning lightweight residual corrections from deployment interactions while keeping the pretrained world model frozen, and study whether dynamics shifts are better corrected inside the latent predictor, at its output, or both.

% \paragraph{Parameter-efficient fine-tuning}
% Parameter-efficient fine-tuning (PEFT) adapts a pretrained network with most of its parameters frozen. LoRA uses trainable low-rank weight updates~\citep{hu2021lora}, bottleneck adapters add nonlinear residual branches to Transformer blocks~\citep{houlsby2019parameter}, and Adapter$^{+}$ scales such a branch and places it after the feed-forward block~\citep{steitz2024adapters}. Another work corrects the network output instead~\citep{zeng2020tossingbot}. As PEFT usually targets task or appearance transfer, it remains open whether a dynamics shift is better corrected inside a latent predictor or after it. Our dual-site adapter occupies both locations.

% \paragraph{Test-time training and adaptation}
% Test-time training and adaptation update a deployed model on unlabeled test data through a self-supervised or entropy-based objective, from image classification under corruption~\citep{sun2020test,wang2020tent} to visual policies that keep learning after deployment~\citep{hansen2020self}, and we apply the same principle to the latent dynamics of a frozen world model.

\paragraph{World model adaptation to dynamics shifts}
Prior work follows three main strategies. \emph{Explicit identification} estimates physical parameters: PIN-WM~\citep{li2025pin} fits inertia, friction, and restitution through a differentiable simulator, while Incremental Few-Shot Adaptation~\citep{baumeister2025incremental} identifies simulator parameters from interaction data. Both require a simulator and suitable parameterization. \emph{Implicit conditioning} infers dynamics from interaction history. DyWA~\citep{lyu2025dywa} conditions its predictor on a learned dynamics embedding, but remains limited to shifts represented during training. \emph{Direct adaptation} uses prediction errors at deployment. AdaJEPA~\citep{wang2026adajepa}, Simulation Distillation~\citep{levy2026simulation}, and AdaWorldPolicy~\citep{yuan2026adaworldpolicy} update pretrained weights, whereas ReDRAW~\citep{lanier2025adapting} and SPREAD~\citep{moon2026spread} keep the backbone frozen but adapt only inside the predictor. In contrast, we keep the pretrained model frozen and learn self-supervised residual corrections both inside the predictor and on its output.

\section{Method}\begin{figure}[t]
    \centering
    \includegraphics[width=1\linewidth]{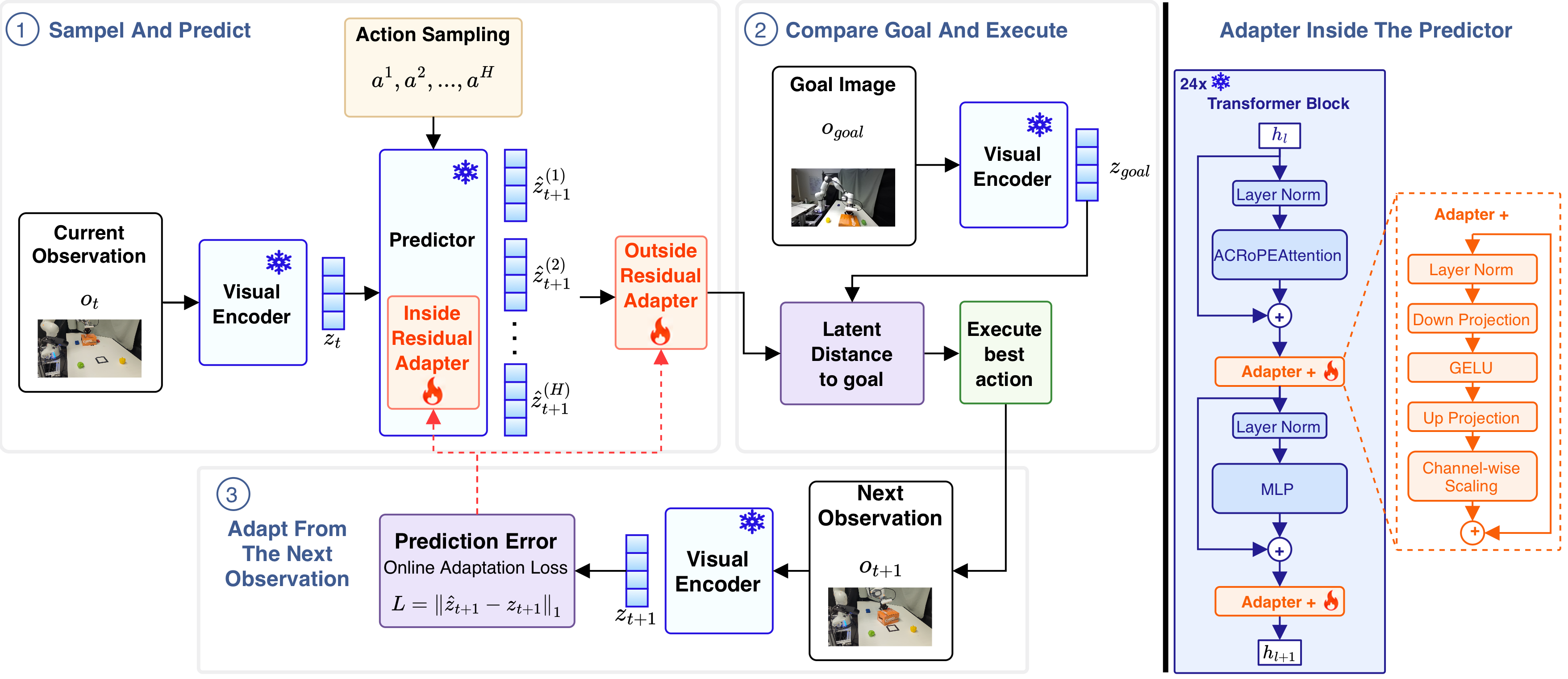}
    \caption{Dual-site latent correction in a frozen world model. The frozen
predictor rolls out sampled actions, corrected inside every block and once more
on its output. The CEM controller ranks the corrected predictions by latent
distance to the encoded goal and executes the best action. The next observation
supplies the target of the self-supervised loss that trains both
residuals.}
    \label{fig:OverviewArchitecture}
\end{figure}
We adapt a pretrained latent world model to the dynamics of a deployment scene
without changing any of its original parameters (Figure~\ref{fig:OverviewArchitecture}). The pipeline alternates
model-predictive control with self-supervised adaptation over three
interaction episodes. Our dual-site adapter corrects the model in two places,
inside its latent predictor and at the predictor output. Both corrections are
frozen once adaptation ends, and the original planning procedure is retained
throughout.

\subsection{Latent world model and planning}

Our base model is DUET-DINO~\citep{nilavadi2026duet}, a variant of V-JEPA 2-AC~\citep{assran2025v}. It predicts action-conditioned latent transitions from a static side camera and a wrist camera. A frozen DINOv3 encoder maps each $224\times224$ image to $14\times14$ patch tokens of dimension 1280, and cross-attention exchanges information bidirectionally between the two views~\citep{simeoni2025dinov3}. Separate side- and wrist-view heads predict the next latent tokens of each view. Each head comprises 24 frame-causal Transformer blocks of width 1024, conditioned on the seven-dimensional end-effector state $\mathbf{s}_t$ and action $\mathbf{a}_t$. The gripper
is held closed throughout, so that last dimension is constant and is neither
planned nor predicted. Writing $\mathbf{z}_t^v$ for the encoded observation of
view $v\in\{\mathrm{s},\mathrm{w}\}$, the one-step prediction is
\begin{equation}
    \widehat{\mathbf{z}}_{t+1}^v
    = \mathrm{LN}\!\left(f_{\theta}^v(\mathbf{z}_t^{\mathrm{s}},
      \mathbf{z}_t^{\mathrm{w}},\mathbf{s}_t,\mathbf{a}_t)\right),
\end{equation}
with $\theta$ the pretrained cross-attention and predictor parameters.

Planning uses a one-step cross-entropy method (CEM) controller \citep{rubinstein1999cross}. At each robot step, it samples actions, predicts the corresponding side and wrist-view latents, and ranks them by latent distance to the encoded goal. After two CEM iterations, the proposal mean is executed. We keep this controller fixed and adapt only the latent predictions it consumes.

% Planning uses a one-step cross-entropy method (CEM) controller \citep{rubinstein1999cross}. At every robot
% step it samples 350 actions from a Gaussian, predicts the resulting
% side- and wrist-view latents, and scores them by mean squared error against
% encoded goal images. After the losses are standardized across candidates,
% translation is ranked by $0.6d_{\mathrm{s}}+0.4d_{\mathrm{w}}$ and rotation by
% $0.2d_{\mathrm{s}}+0.8d_{\mathrm{w}}$, so that each subspace is judged mainly by
% the view that observes it most directly. The ten best candidates per subspace
% give loss-weighted Gaussian statistics for the next
% iteration. After two iterations the proposal mean is executed and planning
% restarts from the new observation. We leave this controller untouched and
% change only the latent predictions it consumes.

\subsection{Dual-site latent correction}

A dynamics shift changes the action-conditioned transition that the pretrained predictor must model. We therefore correct both its internal computation and its final latent prediction, while keeping the encoder, cross-attention blocks, and all pretrained predictor weights frozen.

% A dynamics shift can change both how the predictor processes an action and
% where its final prediction lands in latent space. We therefore correct the
% model at two sites, while the encoder, the cross-attention blocks, and all
% pretrained predictor weights stay frozen.

\textbf{Inside the predictor.}
We insert Adapter$^{+}$ modules at two points in every block of both predictor
heads, after the attention residual and after the completed block~\citep{steitz2024adapters}. For a hidden
token $\mathbf{h}\in\mathbb{R}^{1024}$, each module computes
\begin{equation}
    \mathbf{h}'=\mathbf{h}+A(\mathbf{h}),\qquad
    A(\mathbf{h})=\mathbf{c}\odot
    \mathbf{W}_{\mathrm{up}}\,
    \mathrm{GELU}\!\left(\mathbf{W}_{\mathrm{down}}
    \mathrm{LN}(\mathbf{h})\right),
\end{equation}
with $\mathbf{W}_{\mathrm{down}}\in\mathbb{R}^{128\times1024}$,
$\mathbf{W}_{\mathrm{up}}\in\mathbb{R}^{1024\times128}$ and a learned channel
scale $\mathbf{c}\in\mathbb{R}^{1024}$ that sets the strength of the residual
per feature. Two sites across 24 blocks and two heads give 96 modules in total.
They can reshape the intermediate computation without touching the pretrained
backbone.

\textbf{Outside the predictor.}
A shared token-wise residual MLP then corrects whatever the adapted predictor produces, $\widetilde{\mathbf{z}}_{t+1}^v=\widehat{\mathbf{z}}_{t+1}^v
+R(\widehat{\mathbf{z}}_{t+1}^v)$ for either view. It is
$1280\rightarrow1024\rightarrow1024\rightarrow1280$ with GELU activations and no input normalization, and sees only the predicted token, not the previous latent or the action. Sharing it across views supplies one common output-space correction, while the view-specific internal adapters keep separate capacity.

We zero-initialize the up-projection of every internal adapter and the final layer of the external MLP, so both residuals vanish before the first update and the adapted model starts out equivalent to the frozen base model. Together the internal adapters (25.6M parameters) and the external MLP (3.7M) add 29.2M trainable parameters, 4.53\% of the frozen predictor.

\subsection{Self-supervised online adaptation}

The adapters learn only from prediction errors between predicted and observed transitions. Each episode is subsampled to 4~Hz, encoded once with the frozen encoder, and the resulting layer-normalized latents are cached in a replay buffer. The next encoded observation serves as the target for the one-step objective:
\begin{equation}
    \mathcal{L}_{\mathrm{pred}}
    = \sum_{v\in\{\mathrm{s},\mathrm{w}\}}
      \left\lVert
      \widetilde{\mathbf{z}}_{t+1}^v-\mathbf{z}_{t+1}^v
      \right\rVert_1 .
\end{equation}
Neither rewards nor knowledge of the changed physical parameters are required.

After each of the first three episodes, we update the adapters sequentially in two phases. First, the internal Adapter$^{+}$ modules are trained for 25 steps on batches of eight 8-frame clips using teacher-forced and two-step autoregressive losses over both views. Clips come from the current episode initially and the replay buffer thereafter. We then freeze these adapters, recompute one-step predictions, and train the external MLP for 25 steps on batches of up to 256 transitions with an $\ell_1$ loss. Both phases use AdamW with learning rates $4.343\times10^{-4}$ (inside) and $1.234\times10^{-3}$ (outside).

% After each of the first three episodes we run a two-phase update, in which the
% second phase sees the result of the first. Phase one trains only the internal
% Adapter$^{+}$ modules for 25 gradient steps on batches of eight 8-frame clips,
% under a loss combining teacher-forced predictions with a two-step
% autoregressive term over both views. Its clips come from the newly collected
% episode in the first round and, in later rounds, from the replay buffer holding
% the preceding episodes. Phase two freezes those updated adapters, recomputes
% one-step predictions for every recorded transition, and trains the external MLP
% for 25 gradient steps on batches of up to 256 transitions with an $\ell_1$
% loss. Both phases use AdamW, with learning rates
% $4.343\times10^{-4}$ inside and $1.234\times10^{-3}$ outside.

\section{Evaluation}

We change only the physical dynamics while keeping the task, goal, and planning fixed. Our evaluation addresses four research questions: 
\textbf{(RQ1)} How do physical dynamics shifts affect planning with a pretrained latent world model?, 
\textbf{(RQ2)} How much of the planning can a few-shot adaptation recover?,
\textbf{(RQ3)} How does correction placement affect adaptation?, and \textbf{(RQ4)} What does the learned adaptation capture?

% We change only the physical dynamics to isolate their effects, while keeping the task, goal, and controller fixed. Across simulation and the real robot, we structure our study around four research questions that progressively examine failure, recovery, adaptation location, and the nature of the learned correction.

% \begin{enumerate}[label=\textbf{RQ\arabic*:}, leftmargin=*]
%     \item How do physical dynamics shifts affect pretrained latent world models?
%     \item How far does few-shot calibration/adaptation restore control under a dynamics shift?
%     \item  How does the placement of the correction affect adaptation?
%     \item What does a learned adaptation capture?
% \end{enumerate}

\subsection{Setup}

\textbf{Task and dynamics shifts.} We study goal-directed box pushing, where the robot uses its closed gripper to push a box to a target pose. 
Under \emph{general} dynamics, the box has a centered center of mass and uniform friction between the box and the tabletop, so a forward push moves it toward the goal without rotation. 
We evaluate on two shifts while keeping the task fixed. 
In the \emph{center-of-mass} (CoM) shift, a hidden weight moves the center of mass to the right. 
In the modified \emph{friction}, a cloth under half of the track creates asymmetric friction across the box. 
Note that the scene stays visually unchanged under the CoM shift, whereas the cloth of the modified friction is visible.
% The robot pushes a box across a table into a goal pose without grasping it, so the closed gripper acts as the pushing tool. 
% Under \emph{general} dynamics the box has the center-of-mass in the middle and the whole track has the same surface friction, so a straight push moves the box toward the goal without turning
% it.
% We consider two dynamics shifts while keeping the task unchanged. In the \emph{center-of-mass} shift, a hidden weight inside the box that displaces its center of mass to the right side. In the \emph{friction} shift,  a cloth is placed under one half of the track, causing the two sides of the box to move over surfaces with different friction. The center-of-mass shift leaves the scene visually unchanged, while the cloth under the friction shift is visible.

\begin{figure}[t]
    \centering
    \includegraphics[width=0.8\linewidth]{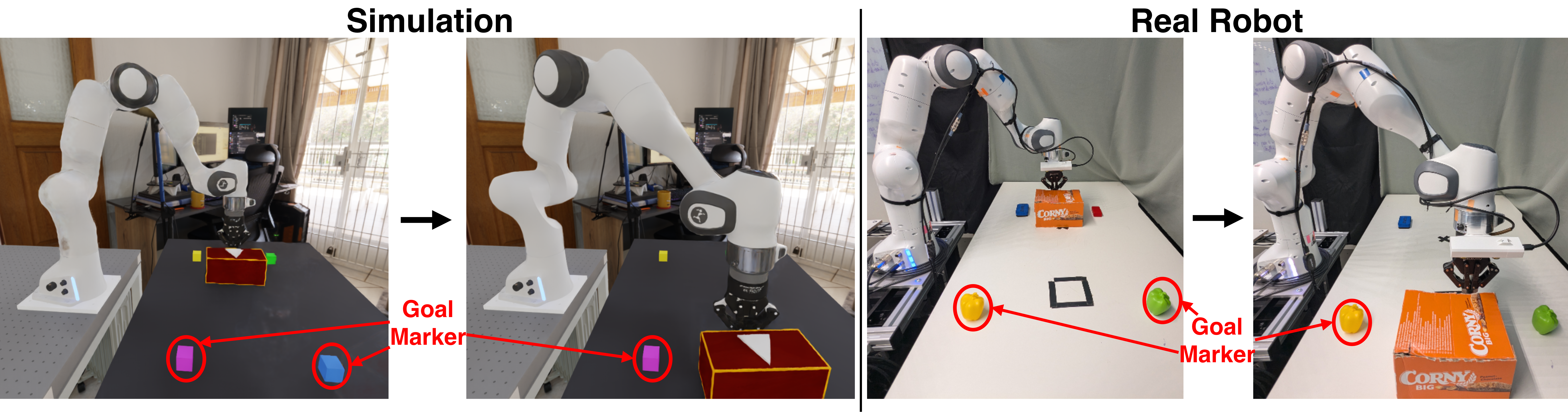}
    \caption{The pushing task in simulation and on the real robot, each shown at the start of an episode and after a successful
push.}
    \label{fig:SimTaskExplanation}
\end{figure}

% \textbf{Platforms}
% \emph{Real robot.} We use a Franka Panda arm in the DROID configuration, with a static side view camera and a second wrist camera \citep{khazatsky2024droid}. Planning and calibration run on a single RTX~4060~Ti. One control step takes roughly 28 seconds end to end, and the three calibration episodes contain only 24 frames in total. Both dynamics shifts are evaluated here.
% \emph{Simulation.} We use RoboLab to reproduce the same tabletop setup with both camera views \citep{yang2026robolab}. Each episode lasts 25 steps, giving 75 recorded frames across the three calibration episodes. We evaluate only the center-of-mass shift, as the split-friction surface did not reproduce faithfully in RoboLab.
\textbf{Platforms.}
\emph{Real robot:} We use a Franka Panda in the DROID configuration with a static side camera and a wrist camera \citep{khazatsky2024droid}. 
Model inferences are run on a single RTX~4060~Ti. 
Each control step takes about $28$ sec end-to-end, and the three adaptation episodes contain only $24$ frames in total. 
We evaluate both dynamics shifts.
\emph{Simulation:} We reproduce the same two-view tabletop setup in the RoboLab simulator~\citep{yang2026robolab}. 
Each episode lasts $25$ steps, yielding $75$ frames across three adaptation episodes. 
We evaluate only the center-of-mass shift because the split-friction surface could not be reproduced faithfully in RoboLab.

\textbf{Protocol.}
We use the same CEM planning in simulation and on the real robot, except for the maximum action norm: $0.25$ in simulation and $0.2$ on the real robot~\citep{rubinstein1999cross}. In simulation, we additionally evaluate a Diffusion Policy~\citep{chi2024diffusionpolicy} as a non-world-model baseline.
An episode is successful if at least half of the box crosses the goal marker, which corresponds to $8$~cm on the real robot and $4.7$~cm in simulation, while its orientation stays within about $25^\circ$ of the starting orientation (see Figure~\ref{fig:SimTaskExplanation}).  
For the architecture comparison in simulation, we report $150$ runs per method across three adaptation seeds, with $50$ runs per seed. 
Each real-robot experiment uses $30$ runs.

\subsection{Results}

\textbf{RQ1: How do physical dynamics shifts affect planning with a pretrained latent world model?} 
As shown in Table~\ref{tab:results_baselines_alt}, under general dynamics, Diffusion Policy achieves 96.0\% success in simulation~\citep{chi2024diffusionpolicy} and the frozen DUET-DINO model reaches 85.3\% in simulation and 86.7\% on the real robot. 
After shifting the box's center of mass, success drops to 0\% for all baselines. After changing friction, the frozen DUET-DINO model reaches only 3.3\% on the real robot.
This shows that changes in object properties lead to changes in physical dynamics, which in turn break planning.

\begin{table}[htbp]
  \centering
  \small
  \caption{Aggregated success rate over $150$ runs under general dynamics and the two dynamics shifts. Diffusion Policy is evaluated in simulation only.}
  \label{tab:results_baselines_alt}
  \begin{tabular}{@{}lccc@{}}
      \toprule
      & \multicolumn{3}{c}{\textbf{Success rate (\%)} $\uparrow$} \\
      \cmidrule(l){2-4}
      \textbf{Method} & \textbf{General} & \textbf{Shifted CoM} & \textbf{Modified friction} \\
      
      \midrule
      
      \rowcolor{gray!15}
      \multicolumn{4}{@{}l}{\textbf{Simulation}} \\
      
      DUET-DINO             & 85.3 & \phantom{0}0.0 & -- \\
      
      Diffusion Policy            & 96.0 & \phantom{0}0.0 & -- \\
      
      \addlinespace
      
      \rowcolor{gray!15}
      \multicolumn{4}{@{}l}{\textbf{Real robot}} \\
      
      DUET-DINO    & 86.7 & \phantom{0}0.0 & \phantom{0}3.3 \\
      
      \bottomrule
  \end{tabular}
\end{table}

\textbf{RQ2: How much of the planning can a few-shot adaptation recover?}
The frozen DUET-DINO model achieves 0.0\% success under the center-of-mass shift and 3.3\% under the modified friction~(Table~\ref{tab:results_baselines_alt}).
Shift-specific full finetuning and LoRA recover some performance (Table~\ref{tab:results_adapters_merged}), but vary substantially across the two shifts. 
In contrast, our shift-specific dual-site adapter reaches 90.0\% success under the center-of-mass shift and 100.0\% under the modified friction. 
Few-shot adaptation therefore recovers most of the lost planning performance across both evaluation setups.

\begin{table}[h]
  \centering
  \footnotesize
    \caption{Success rates in simulation and on the real robot. Simulation results are mean $\pm$ standard deviation over three seeds (50 runs each). The four best-performing adapters are evaluated on the real robot. Modified friction is evaluated only there as it was unreliable in RoboLab.}
  \label{tab:results_adapters_merged}
  \begin{tabular}{@{}lccc@{}}
      \toprule
      & \multicolumn{3}{c}{\textbf{Success rate (\%)} $\uparrow$} \\
      \cmidrule(l){2-4}
      \textbf{Method} & \textbf{Simulation}  & \multicolumn{2}{c}{\textbf{Real robot}} \\
      \cmidrule(lr){2-2}\cmidrule(l){3-4}
       & \textbf{Shifted CoM} 
       & \textbf{Shifted CoM} 
       & \textbf{Modified friction} \\
      \midrule
      
      \rowcolor{gray!15}
      \multicolumn{4}{@{}l}{\textbf{Frozen Base Policy}} \\
      Diffusion Policy         & \phantom{0}0.0 $\pm$ \phantom{0}0.0 & --   & -- \\
      DUET-DINO (frozen) & \phantom{0}0.0 $\pm$ \phantom{0}0.0 & \phantom{0}0.0 & \phantom{0}3.3 \\
      \midrule
      \rowcolor{gray!15}
      \multicolumn{4}{@{}l}{\textbf{W/ Adapters (Inside)}} \\
      Full finetuning          & 28.0 $\pm$ 10.0 & 20.0 & \textbf{100.0} \\
      LoRA                     & 32.7 $\pm$ \phantom{0}9.0 & 30.0 & 66.7 \\
      Adapter$^{+}$            & 18.0 $\pm$ \phantom{0}5.3 & --   & -- \\
      Residual MLP             & 10.7 $\pm$ \phantom{0}9.0 & --   & -- \\
      Bottleneck adapter       & 12.7 $\pm$ 12.2 & --   & -- \\
      \addlinespace
      \rowcolor{gray!15}
      \multicolumn{4}{@{}l}{\textbf{W/ Adapters (Outside)}} \\
      Residual MLP             & 24.0 $\pm$ \phantom{0}3.5 & --   & -- \\
      Channel-wise scaling     & 23.3 $\pm$ \phantom{0}7.6 & --   & -- \\
      \addlinespace
      
      \rowcolor{gray!15}
      \multicolumn{4}{@{}l}{\textbf{W/ Adapters (Combination)}} \\
      Dual-site adapter, joint & 44.0 $\pm$ \phantom{0}6.0 & 80.0 & 90.0 \\
      
      % \midrule
      \rowcolor{green!15}
      \textbf{Dual-site adapter (Ours)} & \textbf{56.7 $\pm$ 13.3} & \textbf{90.0} & \textbf{100.0} \\
      \bottomrule
  \end{tabular}
\end{table}
\textbf{RQ3: How does correction placement affect adaptation?} We compare corrections inside the predictor, on the predicted latent state, and at both sites. 
As shown in Table~\ref{tab:results_adapters_merged}, no single-site method fully recovers planning: the best internal method, LoRA, reaches 32.7\% success, while the best external correction reaches 24.0\%. 
Our dual-site adapter performs substantially better at 56.7\%, with sequential training also outperforming joint training (56.7\% vs.\ 44.0\%). 
This suggests that correcting both the predictor computation and its output is important. 
The two sites also show complementary failure modes: internal adaptation often moves in the correct direction but can be unstable, while the external residual gives more consistent motion but may move in the wrong direction. 
This helps explain the stronger performance of dual-site adaptation.

% To study where the dynamics correction should be applied, we compare adaptation inside the predictor, outside on the predicted latent state, and at both sites. As written in Table \ref{tab:results_baselines_alt} No single-site method fully recovers control, the best internal method, LoRA, reaches 32.7\% success, while the best external correction reaches 24.0\%. Combining the two sites performs substantially better, and our dual-site adapter has reached a success of 56.7\%. Training the two corrections sequentially also outperforms joint training (56.7\% vs.\ 44.0\%). These results suggest that correcting both the predictor computation and its final latent prediction is important for adaptation.

% We also observe different failure modes across correction sites. Internal
% adaptation can produce actions in the correct direction, but the resulting
% behavior is often unstable or jittery across subsequent steps. In contrast,
% the external residual produces more consistent motions, although these
% corrections do not always move the box in the desired direction. This
% complementarity provides a qualitative explanation for the stronger
% performance of the dual-site adapter.

% \begin{itemize}[nosep]
%   \item Finetune adapter: first push is correct but then it does not work anymore correct and like last time made movements that no other/base model made $\rightarrow$ goes up and not forward.
%   \item Outside residual makes consistent movements but not always correct and inside makes correct actions but very inconsistent and jittery.
% \end{itemize}

\textbf{RQ4: What does the learned adaptation capture?}
We analyze the learned correction on 306 held-out episodes (7,650 transitions). The general-dynamics adapter is trained on the general task with the same three-episode self-supervised procedure as our adapter, and it reduces prediction error almost as much as the CoM-shift adapter: it shares 95.3\% of the side-view and 71.7\% of the wrist-view error reduction. However, under the CoM shift it achieves only 6.0\% success, compared with 72.0\% for the CoM-shift adapter (Table~\ref{tab:focused_diagnostics}). Thus, reducing prediction error alone is not sufficient for successful control. The large control gap suggests that the CoM-shift adapter captures additional shift-specific information that is important for control, although our analysis does not reveal the mechanism or physical quantity encoded by this correction. Adaptation also comes at a cost: the CoM-shift adapter reaches only 54.0\% on the general task, compared to 85.3\% for the frozen model. The adapter remains effective under changes in object color, table, marker, and background, but transfers poorly between simulation and the real robot (Table~\ref{tab:focused_diagnostics}), suggesting that it learns a deployment-specific correction rather than a transferable physics model.

\begin{table}[t]
  \centering
  \footnotesize
\caption{Panel (a) ablates dynamics transfer, adapter location, and domain transfer. The arrow separates the training condition from the test condition. Panel (b) varies the scene appearance at test time. The
adapter is trained on the red box in the unchanged scene and never on these variations. All entries are success rates in percent over 50 runs from a single adaptation seed. }
  \label{tab:focused_diagnostics}

  \begin{minipage}[t]{0.53\textwidth}
    \centering
    \textbf{(a) Adapter ablations across dynamics specificity, correction site, and cross-domain transfer.}\\[3pt]
    \begin{tabular}{@{}l@{\hspace{1.8em}}r@{}}
      \toprule
      \textbf{Variant / protocol} & \textbf{Success rate (\%)} $\uparrow$ \\
      \midrule

      \rowcolor{gray!15}
      \multicolumn{2}{@{}l}{\textbf{Dynamics transfer (train $\rightarrow$ test)}} \\
      General dynamics $\rightarrow$ CoM shift & \phantom{0}6.0 \\
      CoM shift $\rightarrow$ CoM shift         & 72.0 \\
      CoM shift $\rightarrow$ nominal task      & 54.0 \\

      \addlinespace
      \rowcolor{gray!15}
      \multicolumn{2}{@{}l}{\textbf{Adapter location}} \\
      Wrist predictor only     & 18.0 \\
      Side-view predictor only & 14.0 \\

      \addlinespace
      \rowcolor{gray!15}
      \multicolumn{2}{@{}l}{\textbf{Domain transfer (train $\rightarrow$ test)}} \\
      Real robot $\rightarrow$ sim (50 steps) & 10.0 \\
      Real robot $\rightarrow$ sim (25 steps) & \phantom{0}0.0 \\
      Sim $\rightarrow$ real robot            & \phantom{0}0.0 \\
      \bottomrule
    \end{tabular}
  \end{minipage}\hfill
  \begin{minipage}[t]{0.43\textwidth}
    \centering
    \textbf{(b) Appearance generalization}\\[3pt]
    \begin{tabular}{@{}l@{\hspace{1.8em}}r@{}}
      \toprule
      \textbf{Variation} & \textbf{Success rate (\%)} $\uparrow$ \\
      \midrule

      \rowcolor{gray!15}
      \multicolumn{2}{@{}l}{\textbf{Red box (seen)}} \\
      Original scene     & 72.0 \\
      New table          & 84.0 \\
      New marker         & 54.0 \\
      New background     & 60.0 \\

      \addlinespace
      \rowcolor{gray!15}
      \multicolumn{2}{@{}l}{\textbf{Turquoise box (unseen)}} \\
      Original scene     & 74.0 \\
      New table          & 62.0 \\
      New marker         & 74.0 \\
      New background     & 60.0 \\
      New table + marker & 54.0 \\
      \bottomrule
    \end{tabular}
  \end{minipage}
\end{table}

\FloatBarrier

\section{Conclusion and Limitations}

We study few-shot adaptation of pretrained latent world models under changes in physical dynamics. We show how changes in center of mass or friction can break control even when the task and controller remain fixed. By keeping the pretrained model frozen and learning residual corrections from only a few observed transitions, our dual-site adaptation recovers most of the lost performance. Correcting both the predictor computation and its output consistently outperforms single-site adaptation. Our analysis further shows that reducing prediction error alone is not sufficient for successful control. The large gap between general- and shifted-dynamics adapters suggests that successful adaptation captures additional shift-specific information, although our analysis does not reveal the underlying mechanism or physical quantity represented by this correction. The adapter remains effective under visual changes but transfers poorly between simulation and the real robot, suggesting that it learns a deployment-specific correction rather than a generally transferable physics model.

Our current study is limited to center-of-mass and friction shifts in a pushing task and learns a separate residual correction for each dynamics setting. The adapted model may also lose performance when returned to general dynamics. Future work will extend this study to broader dynamics shifts, develop mechanisms for retaining and reusing multiple learned corrections, and investigate the mechanistic basis of these corrections and what dynamics-related information they encode.

% Limitations
% - one adapter one dynamic change
% - loss of performance on normal task
% - no learned physics model which is transferable

% Conclusion:
% Different Dynamics breaks the model -> few sampels are enough to bring back the succes rate 
% some part learns dynamics some part learns the domain

\normalsize

\bibliographystyle{plainnat}
\bibliography{references}

\end{document}